\documentclass{article}
\usepackage{spconf,amsmath,graphicx}
\usepackage{amsfonts}
\usepackage{multirow}
\usepackage{array}
\usepackage{float}
\usepackage{adjustbox}
\newcolumntype{C}[1]{>{\centering\arraybackslash}p{#1}}
\usepackage{booktabs} 

\title{XSA-MAD: Cross-modal Semantic Alignment for \\Morphing Attack Detection}
\name{Jie Jin$^{1,2}$, Mahiro Tokumasu$^{2}$, Yu Makino$^{2}$, Masakatsu Nishigaki$^{1,2}$, Tetsushi Ohki$^{1,2,3}$\thanks{This work was supported in part by JSPS KAKENHI JP23H00463, JP23K28085, and JST Moonshot R\&D Grant Number JPMJMS2215.}}
\address{Graduate School of Science and Technology$^{1}$ Shizuoka University$^{2}$, RIKEN AIP$^{3}$}
\begin{document}
%
\maketitle
%

\begin{abstract} 
Morphing attacks pose a serious threat to face recognition systems. However, existing image-based morphing attack detection (MAD) methods often generalize poorly to unseen generation techniques because they rely solely on visual cues. We propose XSA-MAD, a CLIP-based multimodal framework that explicitly models semantic inconsistencies between bona-fide and morphed faces. Morphing concepts are decomposed into four interpretable attributes, including identity, facial geometry, texture, and consistency, and are encoded as structured and attribute-aware textual representations. The image encoder is progressively aligned with this discriminative textual space, resulting in a unified semantic representation that captures generation-invariant and concept-level discrepancies between bona-fide and morph images. Experiments on MAD22 and MorDIFF, following training on SMDD, demonstrate strong generalization across diverse morphing principles. In particular, XSA-MAD achieves an equal error rate of 2.92\% on GAN-based morphs and consistently outperforms existing methods under high-fidelity generative attacks.

\end{abstract}
\section{Introduction}
\label{sec:intro}
\begin{table*}[t]
\centering
\small
\caption{Prompt structure. The top row shows the fixed base context, followed by four attribute-specific descriptions.}
\label{tab:structured_prompt}
\begin{tabular}{p{2.5cm} p{6.5cm} p{6.5cm}}
\toprule
\textbf{Component} & \textbf{Bona-fide} & \textbf{Morph} \\
\midrule
\textit{Base Context} &
A face captured from a live subject. &
A generative morph spoof. \\
\midrule %
Identity &
Person-specific identity without overlap. &
Blends signatures from distinct individuals. \\
Geometry &
Plausible human alignment across all regions. &
Geometric ambiguity across key features. \\
Texture &
Fine-structure texture persists realistically. &
Texture resolution suppressed by blending.\\
Consistency &
All biometric signals support one unified subject. &
Identity mapping appears contradictory. \\
\bottomrule
\end{tabular}
\end{table*}
\begin{figure*}[t]
    \centering
    \includegraphics[width=\textwidth]{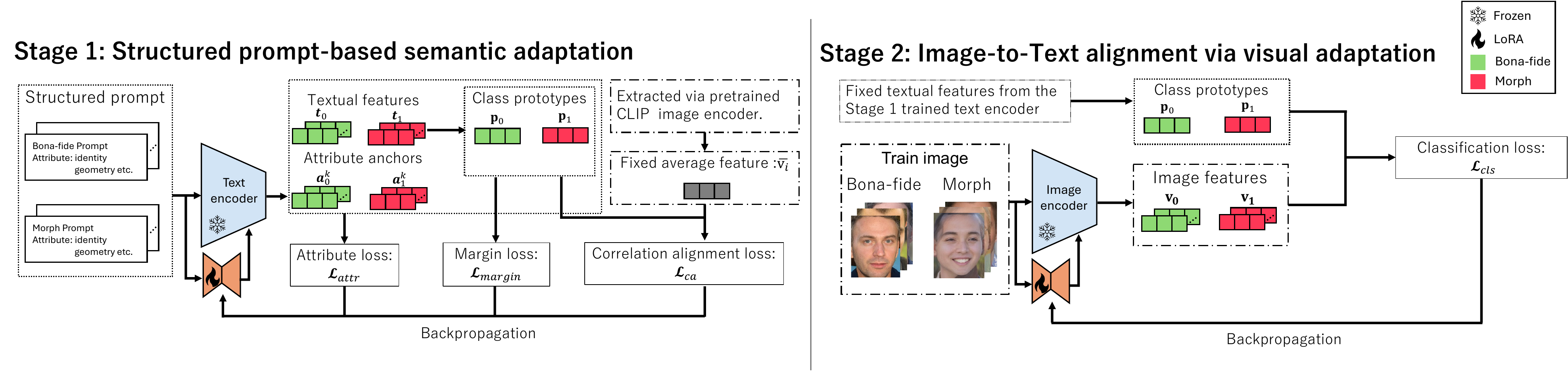}
    \caption{
    Overview of the XSA-MAD framework.
    Stage~1 constructs a discriminative textual semantic space using structured prompts and LoRA.
    Stage~2 aligns visual and textual semantic spaces for MAD.
    }
    \label{fig:overview}
\end{figure*}
Face recognition (FR) systems have been widely deployed as core infrastructure in various real-world applications, including border control, air travel identity verification, and administrative procedures. 
However, with the improved accuracy and widespread adoption of FR systems, morphing attacks that fuse facial characteristics from multiple individuals are increasingly recognized as a major security threat.
Consequently, such attacks have attracted international attention, as the use of morph images as enrollment images can allow identity verification to be passed by multiple individuals, thereby exploiting vulnerabilities in identity verification processes.

Due to the diversification of morph generation techniques, including image-level techniques OpenCV\cite{Mallick2016FaceMorph}, GAN-based methods\cite{zhang2021mipgan}, and diffusion models\cite{damer2023mordiff}, the prevalence of high-quality morph images with minimal visible artifacts has increased significantly. 
This poses substantial challenges for morphing attack detection (MAD), which has therefore emerged as an important research problem.
Most conventional MAD approaches rely on image-based supervised models \cite{damer2021pw,namis2025hybrid,morais2025deep,tapia2025single}, such as CNN and Vision Transformer (ViT), and are therefore strongly biased toward the characteristics of the datasets used for training.
As a result, their performance degrades significantly under different acquisition conditions or when confronted with types of morphing attacks that are not included in the training data\cite{damer2022privacy,zhang2025synmorph,11099381}.
Furthermore, morph images generated by GAN and diffusion models exhibit extremely subtle synthesis artifacts, making visual discrimination between bona-fide and morph images increasingly difficult, and rendering conventional image-based methods progressively less effective.
These challenges stem from the limitation that merely expanding the distribution of image features is insufficient to conceptually capture the differences between bona-fide images and morph images.
These limitations point to the necessity of a framework that explicitly incorporates high-level concepts, such as facial identity and consistency, alongside visual information.

As a potential means of handling such high-level concepts, approaches based on multimodal foundation models have also been explored.
MADation applies the image encoder of CLIP to MAD. However, tuning only the image encoder fails to fully exploit the multimodal nature of CLIP, which relies on the joint representation of images and text\cite{caldeira2025madation, radford2021learning}.
In addition, generic prompts are unable to express morph specific concepts, thereby limiting task-specific discriminative capability.
For example, zero-shot MAD methods utilizing large language models can effectively describe the characteristics of morph images\cite{zhang2025chatgpt}. 
Nevertheless, their detection performance is constrained by the lack of task-specific cross-modal alignment tailored for the MAD domain.


This study proposes a CLIP-based MAD framework that adapts textual and visual semantic spaces through two-stage LoRA optimization~\cite{hu2022lora}. We design structured prompts describing bona-fide and morph images along four key attributes: identity, facial geometry, texture, and consistency. 
In the first stage, the text encoder is adapted to construct discriminative class prototypes from these prompts, while regularizing them to remain consistent with the visual feature distribution. In the second stage, the image encoder is adapted so that image representations are aligned with the established textual prototypes. This stage-wise design separates semantic structure construction from visual alignment, reducing inter-modal interference during optimization. As a result, XSA-MAD forms a unified semantic representation tailored for MAD and improves generalization across diverse attack types.

\section{two-stage Multimodal Optimization}
\label{sec:method}
\subsection{Overview of the Proposed Method}
\label{method_overview}
In this study, we propose XSA-MAD, a two-stage optimization framework to adapt CLIP for MAD, aiming to construct semantic representations that remain invariant to specific morph generation techniques.
Since the original embeddings lack explicit separation, straightforward fine-tuning risks overfitting.
To address this, we employ LoRA to restrict parameter updates to a low-dimensional subspace.

As illustrated in Fig.~\ref{fig:overview}, the framework optimizes the textual and visual semantic spaces sequentially. 
In the first stage, we employ structured prompts decomposing facial characteristics into four attributes to establish a textual semantic space. 
In the second stage, the visual semantic space is optimized using this pre-established textual semantic space as a reference. 
This two-stage optimization allows the model to refine local geometric structures, facilitating the acquisition of abstract discriminative features shared across diverse attacks while avoiding dataset-specific artifacts. 

Although our framework is instantiated using CLIP, the proposed structured prompt design and stage-wise cross-modal alignment are model-agnostic and can, in principle, be applied to other vision–language models with aligned text and image encoders.

\subsection{Structured Prompt Design and Textual Adaptation}
\label{subsec:method_text}
\noindent\textbf{Structured Prompt Construction:}
To capture the complex characteristics of morphing attacks, we utilize a dataset of 80 structured prompts (40 for bona-fide and 40 for morph) generated via ChatGPT 5.1 Thinking. 
The structured prompts were generated using a fixed template consisting of a fixed \textit{Base Context} and four attribute-specific semantic components: \textit{Identity}, \textit{Geometry}, \textit{Texture}, and \textit{Consistency}.
This design relies on structured attribute-level decomposition rather than ChatGPT itself or specific prompt wording. 
The four components are not intended to cover all morphing-related factors; instead, they define compact and interpretable semantic axes, grounded in prior MAD studies, for organizing the textual embedding space.

During prompt generation, constraints were imposed to avoid dataset-specific cues, numerical identifiers, or explicit references to particular morphing algorithms, thereby preventing implicit leakage of dataset or method-specific information. No manual post-editing or class-specific refinement was applied after generation.

As shown in Table~\ref{tab:structured_prompt}, each input sentence is constructed by concatenating the fixed \textit{Base Context} and four attribute-specific semantic conponents.
By feeding these long-form descriptions into the CLIP text encoder, we obtain high-dimensional textual features $\mathbf{t}_c$ that encapsulate multi-faceted semantic information regarding the authenticity or inconsistency of facial images.

\noindent\textbf{Attribute Anchors Definition:}
We introduce eight distinct attribute anchors $\mathbf{a}_{c}^{k}$ ($2 \text{classes} \times 4 \text{attributes}$) to serve as stable reference points in the semantic space.
These anchors are generated by encoding class-specific attribute strings (e.g., ``bonafide\_identity", ``morph\_texture") using the CLIP text encoder.
This ensures that each anchor represents a specific semantic direction tailored to the MAD task.

\noindent\textbf{Class Prototypes:} 
While separation via attribute anchors emphasizes local conceptual differences, it does not inherently ensure that the representative directions of the entire classes are sufficiently distinct. 
To address this, we must maintain an explicit distance between the bona-fide and morph class centers, independent of the attribute-wise directions. 
We therefore define the class prototype $\mathbf{p}_c$ as the mean direction of $N_c$ normalized textual features $\mathbf{t}_i$ belonging to class $c$:
\begin{equation}
\mathbf{p}_c = \frac{1}{N_c} \sum_{i=1}^{N_c} \frac{\mathbf{t}_i}{\|\mathbf{t}_i\|}.
\label{eq:proto}
\end{equation}

\noindent\textbf{Attribute Loss $\mathcal{L}_{\text{attr}}$:}
To enforce a robust discriminative structure, we optimize a unified alignment loss across all attributes $k \in \{\text{Identity}, \allowbreak \text{Geometry}, \allowbreak \text{Texture}, \allowbreak \text{Consistency}\}$. 
We apply a margin-based contrastive constraint to both individual features $\mathbf{t}_{c,i}$ and class prototypes $\mathbf{p}_c$ simultaneously:
\begin{equation} 
\mathcal{L}_{\text{attr}} = \sum_{k, c, \mathbf{x}} [m_\text{attr} + \Delta\cos_{k}(\mathbf{x})]_+ + \sum_{k} [m_\text{attr} + \cos(\mathbf{a}_{c'}^k, \mathbf{a}_c^k)]_+ \label{eq:clean_attr_loss},
\end{equation}
where $\Delta\cos_{k}(\mathbf{x}) = \cos(\mathbf{x}, \mathbf{a}_{c'}^k) - \cos(\mathbf{x}, \mathbf{a}_c^k)$ $\mathbf{x} \in \{\mathbf{t}_{c,i}, \mathbf{p}_c\}$, $c' \neq c$, and $[ \cdot ]_+ = \max(0, \cdot)$. 
The first term pulls both individual prompt features $\mathbf{t}_{c,i}$ and class prototypes $\mathbf{p}_c$ toward their corresponding attribute anchors $\mathbf{a}_c^k$ while pushing them away from opposite class anchors.
The second term explicitly penalizes high similarity between class-specific anchors, ensuring that the semantic reference points for bona-fide and morph classes remain distinct for each attribute.

\noindent\textbf{Margin Loss $\mathcal{L}_{\text{margin}}$:}
To stabilize the global relationship between the bona-fide and morph classes, we introduce a margin-based loss $\mathcal{L}_{\text{margin}}$ based on the cosine similarity between the class prototypes:
\begin{equation}
\mathcal{L}_{\text{margin}} = \max(0, \cos(\mathbf{p}_0, \mathbf{p}_1) + m_{\text{proto}}).
\label{eq:proto-cos}
\end{equation}

The prototype margin $m_{\text{proto}}$ specifies the minimum required separation between class centers. 
This loss acts only when the class prototypes are excessively close, providing to ensure global class separation by penalizing high similarity between the two centers in the semantic space.

\noindent\textbf{Correlation Alignment Loss $\mathcal{L}_{\text{ca}}$:}
Optimizing the textual semantic space in isolation may cause its discriminative structure to diverge from the visual semantic space. 
To prevent this geometric misalignment between modalities, we constrain the class prototypes to remain consistent with the representative direction of the visual semantic space. 
Specifically, let $\bar{\mathbf{v}}_{\text{i}}$ denote the mean image feature across the training dataset, which we treat as the representative direction in the visual semantic space. 
We introduce the correlation alignment loss $\mathcal{L}_{\text{ca}}$ to ensure that the prototypes $\mathbf{p}_0$ and $\mathbf{p}_1$ do not excessively deviate from this direction:
\begin{equation}
\mathcal{L}_{\text{ca}} = \bigl(1 - \cos(\mathbf{p}_0, \bar{\mathbf{v}}_{\text{i}})\bigr) + \bigl(1 - \cos(\mathbf{p}_1, \bar{\mathbf{v}}_{\text{i}})\bigr).
\label{eq:limg}
\end{equation}

This loss ensures that the textual class prototypes do not deviate excessively from the reachable manifold of the visual features, facilitating the subsequent alignment stage.
By incorporating $\mathcal{L}_{\text{ca}}$ into the LoRA-based optimization of the text encoder, the textual semantic space is regularized to maintain modality-wide consistency.

\noindent\textbf{Application of LoRA:}
To update the text encoder based on the defined losses, we apply LoRA to the linear transformations within the Transformer blocks. 
Specifically, LoRA modules target the query ($W_q$), key ($W_k$), and value ($W_v$) projection matrices of the self-attention layers. 
This selective adaptation optimizes the textual semantic space to emphasize the proposed discriminative structure while preserving the foundational knowledge of the pre-trained CLIP model.

\subsection{Image-to-Text Alignment via Visual Adaptation}
\label{subsec:method_image}
\noindent\textbf{Classification Loss $\mathcal{L}_{\text{cls}}$:}
For an input image $i$, the similarity score $z_{i,c}$ for each class is computed as the cosine similarity between the image features $\mathbf{v}_i$ and the class prototypes $\mathbf{p}_c$:
\begin{equation}
z_{i,c} = \cos(\mathbf{v}_i, \mathbf{p}_c).
\label{eq:image-score}
\end{equation}

To enforce a clear separation between classes, we adopt CosFace\cite{wang2018cosface} as our classification objective $\mathcal{L}_{\text{cls}}$:
\begin{equation} 
\mathcal{L}_{\text{cls}} = -\log \frac{\exp\bigl(s \cdot (z_{i,c}-m_{\text{img}})\bigr)}{\exp\bigl(s \cdot (z_{i,c}-m_{\text{img}})\bigr) + \exp\bigl(s \cdot z_{i,1-c}\bigr)}, 
\label{eq:cosface-image} 
\end{equation}
where $s$ is a scaling factor and $m_{\text{img}}$ is the cosine margin. 
This margin intentionally reduces the similarity score of the positive class to widen the decision boundary. 
To backpropagate this loss into the image encoder, LoRA is applied to the $W_q$ and $W_v$ projection matrices of the ViT. 
Through this optimization, the image feature vector $\mathbf{v}_i$ is encouraged to converge toward its corresponding text prototype, ensuring that the visual space is aligned with the discriminative structure established in the textual semantic space.

This multimodal alignment enables the proposed method to detect morphing attacks based on a set of unified concepts, facilitating stable and robust performance across diverse and unknown morphing attacks.
\section{Experiments}
\label{sec:experiments}
We evaluate three core aspects of our framework:
(i) the validity of the two-stage strategy compared to unaligned semantic anchoring (S1-only), unanchored visual alignment (S2-only), and joint training baselines;
(ii) the generalization performance across multiple datasets to assess the detection efficacy against unknown morphing attacks; and 
(iii) the internal structure of the established semantic space, specifically the structural soundness of the textual semantic space and the alignment consistency between image representations and text prototypes.
These joint observations provide a detailed discussion of the quantitative results and the efficacy of the proposed multimodal integration.

\subsection{Datasets}
We use SMDD for training, a privacy-preserving dataset containing 39,730 images (24,735 bona-fide and 14,995 morph) generated via StyleGAN and OpenCV-based morphing.
For evaluation, we use MAD22 and MorDIFF to assess generalization to unseen morph generation methods and data distributions.
MAD22 is organized by morphing generation methods and includes 204 bona-fide images and 4,482 morph images generated using both image-based tools (FaceMorpher, OpenCV, and WebMorph) and GAN-based models (MIPGAN-I/II).
MorDIFF contains 1,000 high-fidelity diffusion-based morphs generated by diffusion autoencoders\cite{preechakul2022diffusion}.
This cross-dataset evaluation tests robustness across diverse generation principles not observed during training.

\subsection{Training Setup}
We adopt CLIP ViT-B/16 and apply LoRA with $r=4$ and $\alpha=8$, while freezing all pretrained weights.

\noindent\textbf{Stage~1:}
We insert LoRA into the text encoder, targeting $W_q$, $W_k$, and $W_v$, with a dropout rate of 0.2.
The inputs are structured prompts comprising 40 bona-fide and 40 morph sentences.
We optimize the text encoder for 100 epochs using AdamW\cite{loshchilovdecoupled} with $lr=10^{-5}$ and $wd=10^{-2}$ to construct a  textual semantic space.

\noindent\textbf{Stage~2:}
We fix the class prototypes and train the CLIP image encoder with LoRA on $W_q$ and $W_v$, using a dropout rate of 0.4 for 2 epochs with balanced mini-batches of size 128.
This limited training duration is sufficient because the encoder only aligns its image features with the class prototypes from Stage~1 while preventing over-adaptation.
Image representations are aligned to these prototypes using an additive cosine margin with $m_{\text{img}}=0.5$ and  $s = 32$.
These LoRA settings, dropout rates, margins, and loss weights were selected through preliminary experiments to balance training stability and detection performance.
Specifically, we set $m_{\text{attr}}=1.0$, $m_{\text{proto}}=1.0$, $\lambda_{\text{attr}}=0.3$, $\lambda_{\text{margin}}=0.6$, and $\lambda_{\text{ca}}=0.6$.

\subsection{Baselines for Optimization}
\label{subsec:baselines}
To validate our two-stage optimization, we compare it with joint-training baselines that simultaneously adapt both encoders. 
These baselines share the same frozen CLIP backbone, LoRA-injected matrices, and structured prompts as XSA-MAD. 
We evaluate three variants: Joint(Text) and Joint(Img), step-matched to Stage~1 and Stage~2 respectively, and Joint(Loss), trained for 40 epochs until loss minimization.

\subsection{Evaluation Metrics}
For semantic-space analysis, we measure the angular distance between embeddings to quantify class separability and cross-modal alignment.
For detection performance, we report ISO/IEC 30107-3 metrics\cite{ISOIEC30107-3-2023}: BPCER, APCER, and EER.
We also report APCER@BPCER at fixed BPCER thresholds of 1\%, 10\%, and 20\%, and BPCER@APCER at the corresponding fixed APCER thresholds.
\section{Results and Discussion}
\label{sec:results}

\subsection{Effectiveness of Stage-wise Optimization}
Table~\ref{tab:strategy_results} demonstrates that our stage-wise approach achieves a superior EER of $6.13\%$. 
S1-only results in a substantial performance degradation with an EER of $43.92\%$, as it lacks cross-modal alignment. 
In contrast, S2-only is limited to an EER of $6.78\%$. These results confirm that XSA-MAD succeeds by establishing robust class prototypes as a foundation before performing visual alignment.

This advantage is most pronounced in high-fidelity generative attacks such as MIPGAN-I and MorDIFF, where we attain EERs of 3.47\% and 1.88\%, significantly surpassing the best joint baseline, Joint (Img), at 5.99\% and 2.48\%. 
Conversely, the smaller gains on traditional morphing methods such as OpenCV and WebMorph reflect a design trade-off: XSA-MAD prioritizes generation-invariant semantic inconsistencies over low-level blending artifacts, aiming for robustness to high-fidelity generative attacks with minimal visible artifacts.
While our focus on high-level concepts effectively captures generative inconsistencies, it is naturally less sensitive to low-level pixel-blending artifacts typical of legacy interpolation.
While Joint (Img) is competitive at 6.70\%, simultaneous training can be unstable, as seen in the 24.11\% EER of Joint (Loss). 
Decoupling optimization into two stages prevents inter-modal interference, ensuring robust anchors for superior generalization.
Although XSA-MAD uses two-stage training, inference requires only a single CLIP-based image encoding with lightweight LoRA adaptation. On an NVIDIA RTX 6000 Ada Generation GPU with a batch size of 1, reflecting practical deployment, XSA-MAD takes 13.05 ms per image with 1.76 GB peak GPU memory.
\begin{table}[!t]
    \caption{Comparison of detection performance between stage-wise learning and joint training baselines on MAD22 and MorDIFF datasets. Best and second-best results are shown in \textbf{bold} and \underline{underlined}, respectively.}
    \centering
    \begin{adjustbox}{max width=\linewidth}
    \begin{tabular}{|c|c|c|c|c|c|c|c|c|c|}
    \hline
    \multicolumn{2}{|c|}{\multirow{2}{*}{Method}} 
    & \multirow{2}{*}{\centering Test data } 
    & \multirow{2}{*}{\centering EER(\%)} 
    & \multicolumn{3}{c|}{APCER (\%) @ BPCER (\%) } 
    & \multicolumn{3}{c|}{BPCER (\%) @ APCER (\%)} \\
    \cline{5-10}
    \multicolumn{1}{|c}{} &&&&
    \multicolumn{1}{C{16mm}|}{1.00} &
    \multicolumn{1}{C{16mm}|}{10.00} &
    \multicolumn{1}{C{16mm}|}{20.00} &
    \multicolumn{1}{C{16mm}|}{1.00} &
    \multicolumn{1}{C{16mm}|}{10.00} &
    \multicolumn{1}{C{16mm}|}{20.00} \\
    \hline
    \multicolumn{2}{|c|}{\multirow{6}[2]{*}{S1-only}} & FaceMorph & 42.64 & 98.70 & 79.68 & 68.27 & 95.59 & 78.92 & 63.24 \\
    \multicolumn{2}{|c|}{} & MIPGAN I  & 35.79 & 100.00 & 89.10 & 73.10 & 78.92 & 57.35 & 44.61 \\
    \multicolumn{2}{|c|}{} & MIPGAN II & 43.14 & 100.00 & 95.40 & 85.79 & 85.29 & 68.63 & 57.84 \\
    \multicolumn{2}{|c|}{} & OpenCV    & 47.95 & 99.09 & 90.14 & 80.79 & 95.59 & 83.33 & 75.00 \\
    \multicolumn{2}{|c|}{} & WebMorph  & 58.27 & 100.00 & 99.20 & 96.00 & 96.57 & 89.71 & 83.33 \\
    \multicolumn{2}{|c|}{}  & MorDIFF  & 35.74 & 99.60 & 83.50 & 65.20 & 83.33 & 59.80 & 48.53 \\
    \cline{3-10}
    \multicolumn{2}{|c|}{}  & Overall & 43.92 & 99.56 & 89.50 & 78.19 & 89.22 & 72.96 & 62.09 \\
    \hline
    \multicolumn{2}{|c|}{\multirow{6}[2]{*}{S2-only}} & FaceMorph & \underline{\smash{3.91}} & 17.92 & \textbf{0.90} & \textbf{0.30} & \textbf{8.82} & \underline{\smash{1.96}} & 0.98 \\
    \multicolumn{2}{|c|}{} & MIPGAN I  & 11.34 & 41.20 & 11.80 & 7.30 & 77.94 & 13.24 & 2.94 \\
    \multicolumn{2}{|c|}{} & MIPGAN II & 5.45 & 25.33 & 4.70 & 2.20 & 45.10 & 2.94 & 1.47 \\
    \multicolumn{2}{|c|}{} & OpenCV    & \textbf{5.29} & \textbf{25.41} & \textbf{3.76} & \textbf{1.63} & \textbf{44.61} & \textbf{2.94} & \textbf{1.96} \\
    \multicolumn{2}{|c|}{} & WebMorph  & \textbf{11.78} & 74.40 & \textbf{15.20} & \textbf{4.40} & \underline{\smash{44.12}} & \textbf{13.24} & \textbf{6.86} \\
    \multicolumn{2}{|c|}{}  & MorDIFF  & 2.92 & 7.00 & 0.90 & \underline{\smash{0.30}} & \underline{\smash{6.86}} & \textbf{0.00} & \textbf{0.00} \\
    \cline{3-10}
    \multicolumn{2}{|c|}{}  & Overall & 6.78 & 31.87 & \underline{\smash{6.21}} & \underline{\smash{2.69}} & 37.91 & \underline{\smash{5.72}} & \textbf{2.37} \\    
    \hline
    \multicolumn{2}{|c|}{\multirow{6}[2]{*}{Joint (Text)}} & FaceMorph & 7.92 & 18.70 & 5.70 & 2.40 & 28.92 & 4.41 & 0.98 \\
    \multicolumn{2}{|c|}{} & MIPGAN I  & 36.73 & 78.00 & 64.60 & 52.10 & 90.69 & 68.63 & 54.41 \\
    \multicolumn{2}{|c|}{} & MIPGAN II & 26.84 & 61.16 & 45.35 & 33.83 & 86.27 & 59.80 & 38.24 \\
    \multicolumn{2}{|c|}{} & OpenCV    & 17.06 & 37.20 & 21.14 & 11.89 & 56.86 & 22.55 & 12.25 \\
    \multicolumn{2}{|c|}{} & WebMorph  & 37.77 & 86.20 & 71.40 & 57.00 & 98.53 & 76.96 & 64.71 \\
    \multicolumn{2}{|c|}{}  & MorDIFF  & 9.80 & 21.00 & 9.30 & 4.50 & 48.04 & 9.31 & \underline{\smash{1.47}}\\
    \cline{3-10}
    \multicolumn{2}{|c|}{}  & Overall & 22.69 & 50.38 & 36.25 & 26.95 & 68.22 & 40.28 & 28.68 \\
    \hline
    \multicolumn{2}{|c|}{\multirow{6}[2]{*}{Joint (Img)}} & FaceMorph & \textbf{2.82} & \textbf{2.70} & \underline{\smash{1.60}} & 1.20 & 27.45 & \textbf{0.00} & \textbf{0.00} \\
    \multicolumn{2}{|c|}{} & MIPGAN I  & \underline{\smash{5.99}} & \underline{\smash{26.00}} & \textbf{2.10} & \underline{\smash{1.60}} & \underline{\smash{31.86}} & \underline{\smash{2.94}} & \underline{\smash{1.96}} \\
    \multicolumn{2}{|c|}{} & MIPGAN II & \underline{\smash{3.86}} & \underline{\smash{15.12}} & \textbf{1.30} & \underline{\smash{1.00}} & \underline{\smash{20.59}} & \underline{\smash{1.96}} & \underline{\smash{0.98}} \\
    \multicolumn{2}{|c|}{} & OpenCV  & \underline{\smash{7.89}}  & 29.37 & \underline{\smash{6.81}} & 4.37  & 58.82 & \underline{\smash{5.88}} & \textbf{1.96} \\
    \multicolumn{2}{|c|}{} & WebMorph & 17.18 & \underline{\smash{73.00}} & \underline{\smash{20.20}} & 14.60 & 72.55 & 28.92 & \underline{\smash{11.27}} \\
    \multicolumn{2}{|c|}{}  & MorDIFF & \underline{\smash{2.48}} & \underline{\smash{6.90}} & \underline{\smash{0.70}}  & 0.50 & \underline{\smash{6.86}} & \underline{\smash{0.98}}  &  \textbf{0.00} \\
    \cline{3-10}
    \multicolumn{2}{|c|}{}  & Overall & \underline{\smash{6.70}} & \underline{\smash{25.51}} & \textbf{5.45} & 3.88 & \underline{\smash{36.36}} & 6.78 & 2.70 \\
    \hline
    \multicolumn{2}{|c|}{\multirow{6}[2]{*}{Joint (Loss)}} & FaceMorph & 8.32 & 20.80 & 6.60 & 2.80 & 30.39 & 5.39 & 1.47 \\
    \multicolumn{2}{|c|}{} & MIPGAN I  & 40.15 & 83.20 & 73.10 & 58.50 & 93.14 & 75.49 & 64.71 \\
    \multicolumn{2}{|c|}{} & MIPGAN II & 30.91 & 68.07 & 53.75 & 39.74 & 91.67 & 68.14 & 47.06 \\
    \multicolumn{2}{|c|}{} & OpenCV    & 16.67 & 37.40 & 21.04 & 10.57 & 56.86 & 21.57 & 12.25 \\
    \multicolumn{2}{|c|}{} & WebMorph  & 36.78 & 85.80 & 71.20 & 53.80 & 98.53 & 75.49 & 60.29  \\
    \multicolumn{2}{|c|}{} & MorDIFF   & 11.83 & 28.30 & 13.00 & 5.40 & 57.35 & 14.71 & 4.41 \\
    \cline{3-10}
    \multicolumn{2}{|c|}{}  & Overall & 24.11 & 53.93 & 39.78 & 28.47 & 71.32 & 43.46 & 31.70 \\
    \hline
    \hline
    \multicolumn{2}{|c|}{\multirow{6}[2]{*}{XSA-MAD (Ours)}} & FaceMorph & 4.41 & \underline{\smash{15.72}} & 2.90 & \underline{\smash{0.40}} & \underline{\smash{14.71}} & 2.45 & \underline{\smash{0.49}} \\
    \multicolumn{2}{|c|}{} & MIPGAN I  & \textbf{3.47} & \textbf{11.20} & \underline{\smash{2.20}} & \textbf{0.60} & \textbf{14.22} & \textbf{1.47} & \textbf{0.49} \\
    \multicolumn{2}{|c|}{} & MIPGAN II & \textbf{2.92} & \textbf{8.31} & \underline{\smash{1.50}} & \textbf{0.40} & \textbf{10.78} & \textbf{0.49} & \textbf{0.00} \\
    \multicolumn{2}{|c|}{} & OpenCV  & 10.78  & \underline{\smash{27.54}} & 11.89  & \underline{\smash{3.86}} & \underline{\smash{46.57}} & 11.27 & \underline{\smash{2.45}} \\
    \multicolumn{2}{|c|}{} & WebMorph & \underline{\smash{13.32}} & \textbf{61.00} & 26.20 & \underline{\smash{5.40}} & \textbf{43.14} & \underline{\smash{16.67}} & \underline{\smash{11.27}} \\
    \multicolumn{2}{|c|}{}  & MorDIFF & \textbf{1.88}  & \textbf{3.00} & \textbf{0.40}  & \textbf{0.10}  & \textbf{2.94} & \textbf{0.00}  &  \textbf{0.00} \\
    \cline{3-10}
    \multicolumn{2}{|c|}{}  & Overall & \textbf{6.13} & \textbf{21.13} & 7.52 & \textbf{1.79} & \textbf{22.06} & \textbf{5.39} & \underline{\smash{2.45}} \\
    \hline
    \end{tabular}
\end{adjustbox}
  \label{tab:strategy_results}
\end{table}

\subsection{Generalization and Comparison with Prior Methods}
Table~\ref{tab:addlabel} summarizes the comparison between XSA-MAD and prior MAD approaches. 
XSA-MAD remains competitive on conventional image-level morphs while providing clear advantages on advanced generative attacks. 
Specifically, it achieves the lowest EERs on MIPGAN-I and MIPGAN-II at $3.47\%$ and $2.92\%$ respectively, outperforming all compared methods. 
On MorDIFF, XSA-MAD attains an EER of $1.88\%$ and maintains low error rates under strict operating points, such as $0.10\%$ APCER at BPCER=$20\%$ and $0.00\%$ BPCER at APCER=$10\%$ and $20\%$. 

These results confirm that the multimodal space remains discriminative even for high-fidelity morphs with subtle artifacts. 
Overall scores are averaged across all test sets, using only MAD22 for methods lacking MorDIFF results. 
Under this protocol, XSA-MAD consistently yields lower average error rates, confirming robust generalization across fundamentally different generation principles.
\begin{table}[!t]
    \caption{Comparison with prior MAD methods trained on SMDD and evaluated on MAD22 and MorDIFF. Unreported values are shown as "-". Best and second-best results are in \textbf{bold} and \underline{underlined}, respectively.}
    \begin{adjustbox}{max width=0.99\linewidth}
        \begin{tabular}{|c|c|c|c|c|c|c|c|c|c|}
        \hline
        \multicolumn{2}{|c|}{\multirow{2}{*}{Method}} 
        & \multirow{2}{*}{\centering Test data } 
        & \multirow{2}{*}{\centering EER(\%)} 
        & \multicolumn{3}{c|}{APCER (\%) @ BPCER (\%) } 
        & \multicolumn{3}{c|}{BPCER (\%) @ APCER (\%)} \\
        \cline{5-10}
        \multicolumn{1}{|c}{} &&&&
        \multicolumn{1}{C{16mm}|}{1.00} &
        \multicolumn{1}{C{16mm}|}{10.00} &
        \multicolumn{1}{C{16mm}|}{20.00} &
        \multicolumn{1}{C{16mm}|}{1.00} &
        \multicolumn{1}{C{16mm}|}{10.00} &
        \multicolumn{1}{C{16mm}|}{20.00} \\
        \hline
        \multicolumn{2}{|c|}{\multirow{6}[2]{*}{Inception-MAD\cite{damer2023mordiff, ramachandra2019detecting}}} & FaceMorph & \textbf{0.00} & 1.70  & \textbf{0.00} & \textbf{0.00} & -     & -     & - \\
        \multicolumn{2}{|c|}{} & MIPGAN I  & 10.90 & 50.90 & \underline{\smash{13.70}} & 5.70  & -     & -     & - \\
        \multicolumn{2}{|c|}{} & MIPGAN II & 16.22 & 82.48 & 25.83 & 11.41 & -     & -     & - \\
        \multicolumn{2}{|c|}{} & OpenCV  & 7.52  & 28.66 & \underline{\smash{5.49}}  & 3.05  & -     & -     & - \\
        \multicolumn{2}{|c|}{} & WebMorph & 18.00 & 85.20 & 27.40 & 13.40 & -     & -     & - \\
        \multicolumn{2}{|c|}{} & MorDIFF & 5.30  & 17.20 & 3.50  & 2.50  & -     & -     & - \\
        \cline{3-10}
        \multicolumn{1}{|c}{} && Overall & 9.66 & 44.36 & \underline{\smash{12.65}} & \underline{\smash{6.01}} & - & - & - \\
        \hline
        \multicolumn{2}{|c|}{\multirow{6}[2]{*}{Con-Text Net A\cite{10007950}}} & FaceMorph & \textbf{0.00} & 99.90 & \textbf{0.00} & \textbf{0.00} & 100.00 & \textbf{0.00} & \textbf{0.00} \\
        \multicolumn{2}{|c|}{} & MIPGAN I  & 12.30 & \underline{\smash{41.90}} & 14.10 & 8.10  & \underline{\smash{59.31}} & \underline{\smash{16.18}} & \underline{\smash{6.37}} \\
        \multicolumn{2}{|c|}{} & MIPGAN II & 12.91 & 43.44 & 14.51 & 8.61  & \underline{\smash{59.31}} & 19.61 & 5.88 \\
        \multicolumn{2}{|c|}{} & OpenCV  & 17.48 & 70.93 & 26.52 & 15.75 & 74.02 & 32.84 & 16.18 \\
        \multicolumn{2}{|c|}{} & WebMorph & 26.20 & 89.20 & 45.60 & 31.00 & 93.14 & 48.53 & 31.86 \\
        \multicolumn{2}{|c|}{} & MorDIFF & -     & -     & -     & -     & -     & -     & - \\
        \cline{3-10}
        \multicolumn{1}{|c}{} && Overall & 13.79 & 69.07 & 20.15 & 12.69 & 77.16 & 23.43 & 12.06 \\
        \hline
        \multicolumn{2}{|c|}{\multirow{6}[2]{*}{Xception\cite{10007950, ivanovska2022face}}} & FaceMorph & 0.60  & 0.50  & \textbf{0.00} & \textbf{0.00} & 1.47  & \underline{\smash{1.47}}  & 1.47 \\
        \multicolumn{2}{|c|}{} & MIPGAN I  & 36.90 & 97.90 & 80.40 & 57.40 & 86.27 & 57.35 & 49.02 \\
        \multicolumn{2}{|c|}{} & MIPGAN II & 44.54 & 99.50 & 92.49 & 77.08 & 90.20 & 67.65 & 56.86 \\
        \multicolumn{2}{|c|}{} & OpenCV  & 7.32  & \underline{\smash{21.75}} & 6.61  & 2.54  & \underline{\smash{35.29}} & \underline{\smash{4.90}}  & \underline{\smash{1.47}} \\
        \multicolumn{2}{|c|}{} & WebMorph & 14.60 & \underline{\smash{49.40}} & 23.00 & 10.80 & 53.92 & 21.57 & 11.76 \\
        \multicolumn{2}{|c|}{} & MorDIFF & -     & -     & -     & -     & -     & -     & - \\
        \cline{3-10}
        \multicolumn{1}{|c}{} && Overall & 20.79 & 53.81 & 40.50 & 29.56 & 53.43 & 30.59 & 24.17 \\
        \hline
        \multicolumn{2}{|c|}{\multirow{6}[2]{*}{D-FW-MixFaceNet\cite{rachalwar2023depth}}} & FaceMorph & \underline{\smash{0.10}}  & -     & -     & \textbf{0.00} & -     & -     & - \\
        \multicolumn{2}{|c|}{} & MIPGAN I  & \underline{\smash{6.70}} & -     & -     & \underline{\smash{1.20}} & -     & -     & - \\
        \multicolumn{2}{|c|}{} & MIPGAN II & \underline{\smash{6.61}} & -     & -     & \underline{\smash{1.00}} & -     & -     & - \\
        \multicolumn{2}{|c|}{} & OpenCV  & 13.72 & -     & -     & 9.04  & -     & -     & - \\
        \multicolumn{2}{|c|}{} & WebMorph & 10.80 & -     & -     & 7.40  & -     & -     & - \\
        \multicolumn{2}{|c|}{} & MorDIFF & -     & -     & -     & -     & -     & -     & - \\
        \cline{3-10}
        \multicolumn{1}{|c}{} && Overall & 7.59 & - & - & \underline{\smash{3.73}} & - & - & - \\
        \hline
        \multicolumn{2}{|c|}{\multirow{6}[2]{*}{D-FW-CDCN\cite{rachalwar2023depth}}} & FaceMorph & \textbf{0.00} & -     & -     & 44.10 & -     & -     & - \\
        \multicolumn{2}{|c|}{} & MIPGAN I  & 11.90 & -     & -     & 3.80  & -     & -     & - \\
        \multicolumn{2}{|c|}{} & MIPGAN II & 14.11 & -     & -     & 8.51  & -     & -     & - \\
        \multicolumn{2}{|c|}{} & OpenCV  & \textbf{0.30} & -     & -     & \textbf{0.00} & -     & -     & - \\
        \multicolumn{2}{|c|}{} & WebMorph & \textbf{0.00} & -     & -     & 64.00 & -     & -     & - \\
        \multicolumn{2}{|c|}{} & MorDIFF & -     & -     & -     & -     & -     & -     & - \\
        \cline{3-10}
        \multicolumn{1}{|c}{} && Overall & \textbf{5.26} & - & - & 24.08 & - & - & - \\
        \hline
        \multirow{12}[4]{*}{MADation\cite{caldeira2025madation}} & \multirow{6}[2]{*}{ViT-B} & FaceMorph & \textbf{0.00} & \textbf{0.00} & \textbf{0.00} & \textbf{0.00} & \textbf{0.00} & \textbf{0.00} & \textbf{0.00} \\
              &       & MIPGAN I  & 33.37 & 82.97 & 55.18 & 43.92 & 94.12 & 72.55 & 52.94 \\
              &       & MIPGAN II & 22.21 & 79.98 & 34.66 & 24.30 & 84.80 & 47.55 & 26.47 \\
              &       & OpenCV  & \underline{\smash{3.85}}  & \textbf{11.64} & \textbf{1.82} & \underline{\smash{1.11}}  & \textbf{23.53} & \textbf{0.98} & \textbf{0.00} \\
              &       & WebMorph & 10.80 & 60.00 & \underline{\smash{11.40}} & \underline{\smash{5.00}}  &51.47 & \underline{\smash{11.76}} & \underline{\smash{4.41}} \\
              &       & MorDIFF & \textbf{1.10} & \textbf{1.60} & \textbf{0.00} & \textbf{0.00} & \textbf{1.94} & \textbf{0.00} & \textbf{0.00} \\
              \cline{3-10}
              &       & Overall & 11.89 & 39.37 & 17.18 & 12.39 & \underline{\smash{42.64}} & \underline{\smash{22.14}} & 13.97 \\
    \cline{2-10}& \multirow{6}[2]{*}{ViT-L} & FaceMorph & 0.40  & \underline{\smash{0.40}}  & \textbf{0.00} & \textbf{0.00} & \underline{\smash{0.49}}  & \textbf{0.00} & \textbf{0.00} \\
              &       & MIPGAN I  & 20.32 & 55.88 & 29.08 & 20.32 & 79.41 & 35.78 & 15.69 \\
              &       & MIPGAN II & 9.06  & \underline{\smash{19.42}} & \underline{\smash{9.06}} & 5.58 & 100.00 & \underline{\smash{5.39}} & \underline{\smash{0.98}} \\
              &       & OpenCV  & 19.26 & 48.40 & 24.45 & 19.26 & 84.47 & 34.95 & 15.53 \\
              &       & WebMorph & \underline{\smash{2.23}}  & \textbf{3.74} & \textbf{1.32} & \textbf{0.71} & \textbf{15.69} & \textbf{0.00} & \textbf{0.00} \\
              &       & MorDIFF & 20.40 & 47.60 & 20.40 & 20.40 & 82.35 & \underline{\smash{37.25}} & \underline{\smash{13.24}} \\
              \cline{3-10}
              &       & Overall & 11.95 & \underline{\smash{29.24}} & 14.05 & 11.05 & 60.40 & 28.90 & \underline{\smash{7.57}} \\
        \hline
        \hline
        \multicolumn{2}{|c|}{\multirow{6}[2]{*}{XSA-MAD (Ours)}} & FaceMorph & 4.41 & 15.72 & \underline{\smash{2.90}} & \underline{\smash{0.40}} & 14.71 & 2.45 & \underline{\smash{0.49}} \\
        \multicolumn{2}{|c|}{} & MIPGAN I  & \textbf{3.47} & \textbf{11.20} & \textbf{2.20} & \textbf{0.60} & \textbf{14.22} & \textbf{1.47} & \textbf{0.49} \\
        \multicolumn{2}{|c|}{} & MIPGAN II & \textbf{2.92} & \textbf{8.31} & \textbf{1.50} & \textbf{0.40} & \textbf{10.78} & \textbf{0.49} & \textbf{0.00} \\
        \multicolumn{2}{|c|}{} & OpenCV  & 10.78  & 27.54 & 11.89  & 3.86  & 46.57 & 11.27 & 2.45 \\
        \multicolumn{2}{|c|}{} & WebMorph & 13.32 & 61.00 & 26.20 & 5.40 & \underline{\smash{43.14}} & 16.67 & 11.27 \\
        \multicolumn{2}{|c|}{}  & MorDIFF & \underline{\smash{1.88}}  & \underline{\smash{3.00}} & \underline{\smash{0.40}}  & \underline{\smash{0.10}}  & \underline{\smash{2.94}} & \textbf{0.00}  &  \textbf{0.00} \\
        \cline{3-10}
        \multicolumn{2}{|c|}{}  & Overall & \underline{\smash{6.13}} & \textbf{21.13} & \textbf{7.52} & \textbf{1.79} & \textbf{22.06} & \textbf{5.39} & \textbf{2.45} \\
        \hline
        \end{tabular}
        \end{adjustbox}
      \label{tab:addlabel}
\end{table}

\subsection{Semantic Space and Alignment Analysis}
\noindent\textbf{Textual Semantic Space Separation:}
Default prototypes are entangled with $0.23$ separation. 
Despite $0.33$ intra-class coherence, this lacks sufficient separation for MAD. 
Stage~1 optimization increases prototype distance to $2.13$ while maintaining $0.43$ coherence. 
Structural changes, distributions, and attribute ablations are in the Supplementary Material.

\noindent\textbf{Visual--Textual Alignment:}
Alignment consistency is measured using within-class and cross-class angular distances. 
Class-mean image features exhibit high alignment with low within-class distances of $0.64$ for bona-fide and $0.72$ for morph, while cross-class distances are substantially larger at $1.94$ and $1.81$. 
This distance validates the mapping of visual features into the Stage~1 semantic structure. 
Anchoring features to these distinct prototypes prevents noise-overfitting, ensuring generalization across diverse attacks.
\section{Conclusion}
\label{sec:conclusion}
This paper presented XSA-MAD, a two-stage framework adapting CLIP for robust morphing attack detection. 
By decoupling attribute-based textual optimization from LoRA-driven visual alignment, we achieved a unified representation that preserves semantic integrity while mitigating modal interference. 
Evaluations on MAD22 and MorDIFF confirm that XSA-MAD significantly outperforms prior benchmarks, especially against GAN and diffusion-based attacks. 
Ultimately, XSA-MAD establishes a stable, scalable foundation for forensic analysis against evolving generative threats.

\vfill\pagebreak

\bibliographystyle{IEEEbib}
\bibliography{strings,refs}

\end{document}